\documentclass[runningheads]{llncs}
\usepackage{graphicx}
\usepackage[english]{babel}
\usepackage[utf8]{inputenc}
\usepackage{amsmath}
\usepackage{csquotes}
\usepackage[style=authoryear-ibid,backend=biber]{biblatex}
\begin{document}
\title{GenScan: A Generative Method for Populating Parametric 3D Scan Datasets}
%
%
\author{Mohammad Keshavarzi
\and
Oladapo Afolabi
\and
Luisa Caldas
\and
Allen Y. Yang
\and
Avideh Zakhor
}
\institute{University of California, Berkeley}

\authorrunning{Keshavarzi et al.}
%

%
\maketitle              

\begin{abstract}
The availability of rich 3D datasets corresponding to the geometrical complexity of the built environments is considered an ongoing challenge for 3D deep learning methodologies. To address this challenge, we introduce GenScan, a generative system that populates synthetic 3D scan datasets in a parametric fashion. The system takes an existing captured 3D scan as an input and outputs alternative variations of the building layout including walls, doors, and furniture with corresponding textures. GenScan is a fully automated system that can also be manually controlled by a user through an assigned user interface. Our proposed system utilizes a combination of a hybrid deep neural network and a parametrizer module to extract and transform elements of a given 3D scan. GenScan takes advantage of style transfer techniques to generate new textures for the generated scenes. We believe our system would facilitate data augmentation to expand the currently limited 3D geometry datasets commonly used in 3D computer vision, generative design, and general 3D deep learning tasks.

\begin{figure}
  \centering
  \includegraphics[width=\textwidth]{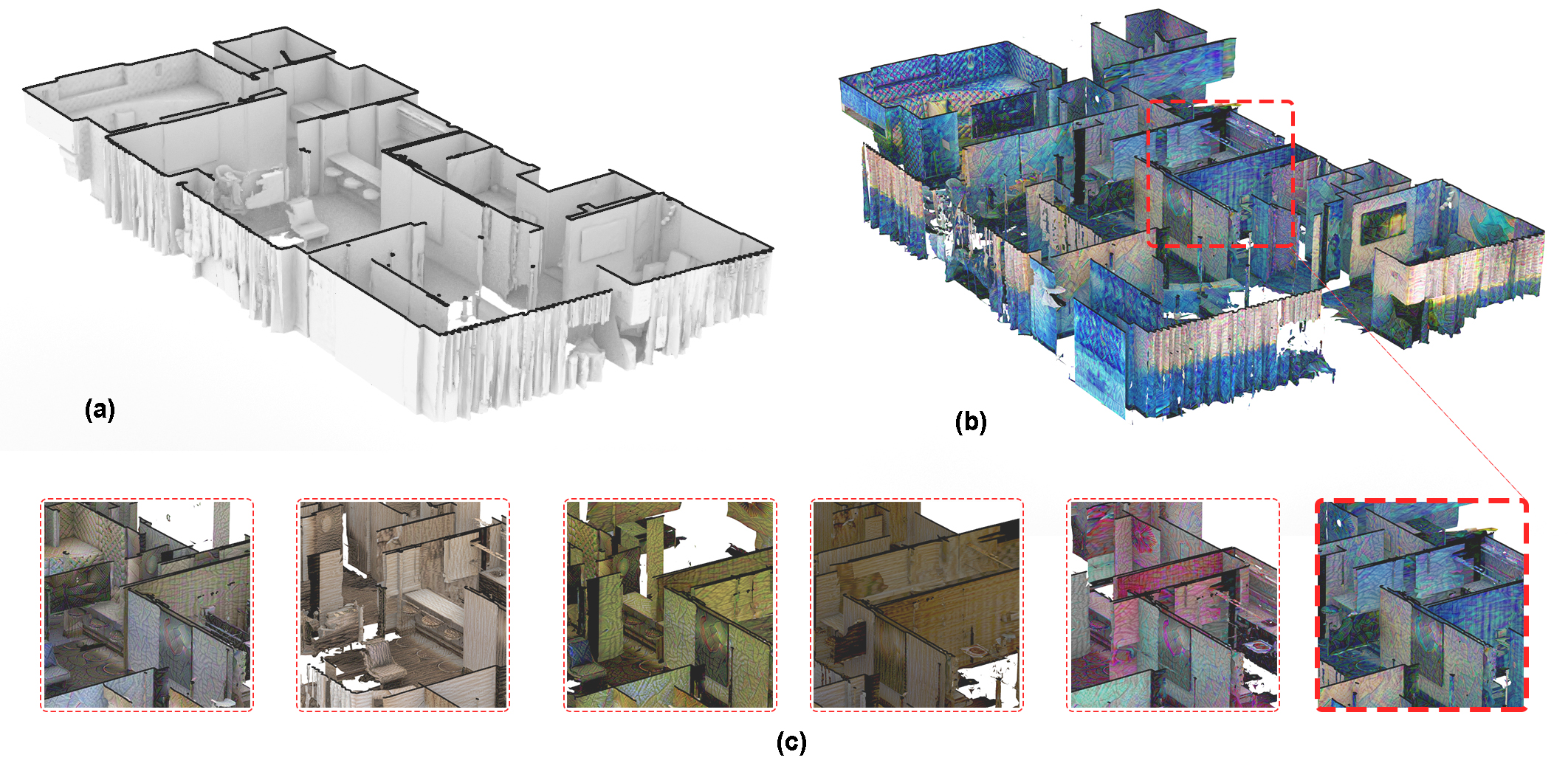}
  \caption{GenScan takes an existing captured 3D scan (a) as an input and outputs alternative parametric variations of the building layout (b) including walls, doors, and furniture with  (c) new generated textures.}
  \label{fig:teaser}
\end{figure}

\keywords{Computational Geometry\and
Generative Modeling
\and
3D Manipulation
\and
Texture Synthesis 
}
\end{abstract}

\section{Introduction}

The utilization of 3D deep learning techniques has been widely explored in cross-disciplinary fields of architecture, computer graphics, and computer vision. For tasks such as synthesizing new environments, semantic segmentation \parencite{Armeni_Sener_Zamir_Jiang_Brilakis_Fischer_Savarese_2016,McCormac_Handa_Davison_Leutenegger_2017,Qi_Su_Niessner_Dai_Yan_Guibas_2016}, object recognition \parencite{Qi_Su_Niessner_Dai_Yan_Guibas_2016}, and 3D reconstruction \parencite{Guo_Zou_Hoiem_2015,Song_Yu_Zeng_Chang_Savva_Funkhouser_2017}, integrating 3D deep learning methodologies have brought a promising direction in the state-of-the-art research. However, like many other learning approaches, the success of this approach is highly dependent on the availability of the appropriate datasets. In contrast to 2D image recognition tasks, where training labeled datasets are available in large quantities, 3D indoor datasets are limited to only a small number of open-source datasets. Capturing 3D geometry is seen to be much more expensive than capturing 2D data in terms of both hardware and human resources

3D data for training resources for computer vision tasks can be found in two general categories (a) real-world captured data and (b) synthetic data. The first approach involves scanning RGB-D data using high end capturing systems or commodity-based sensors. To this extent, a number of open source datasets are available with various scales and capture qualities. The ETH3D dataset contains a limited number of indoor scans \parencite{schops2017multi}, and its purpose is for multi-view stereo rather than 3D point-cloud processing. The ScanNet dataset \parencite{dai2017scannet} and the SUN RGB-D \parencite{song2015sun} dataset capture a variety of indoor scenes with added semantic layers. However, most of their scans contain only one or two rooms, not suitable for larger scale layout reconstruction problem. Matterport3D \parencite{Chang2018} provides high quality panorama RGB-D image sets for 90 luxurious houses captures by the Matterport camera. The 2D-3D-S dataset\parencite{armeni2017joint} provides large-scale indoor scans of office spaces by using the same Matterport camera. 

\begin{figure}
  \centering
  \includegraphics[width=\columnwidth]{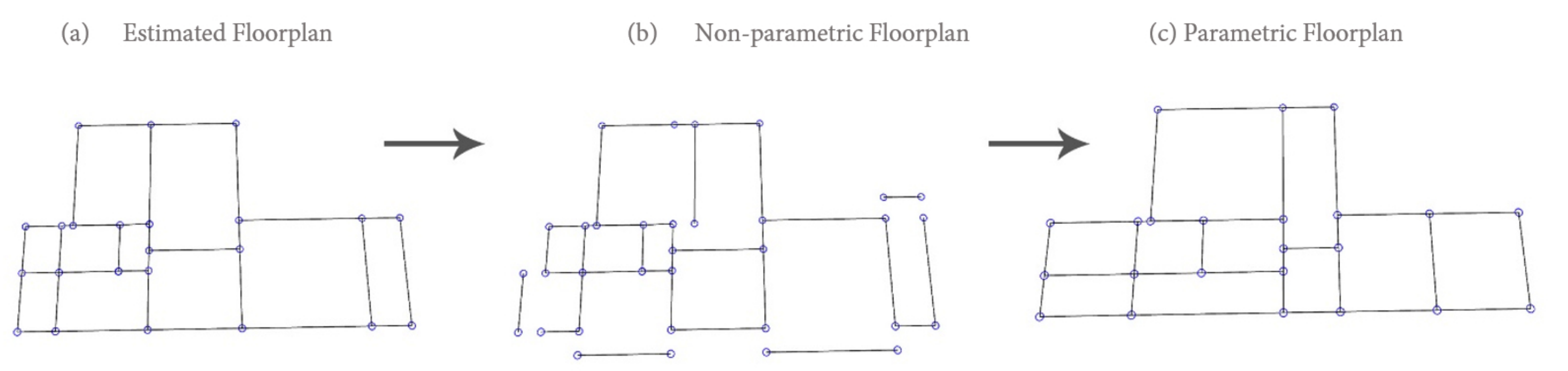}
  \caption{Applying individual transformations to wall segments results in the inconsistency of the output layout (b). Using the Parametrizer module we avoid unwanted voids and opening in the building's walls\label{fig:parameterizer}}
\end{figure}

The second approach is to utilize synthetic 3D data of building layouts and indoor scenes, which has also been recently produced in mass numbers to fill the void of rich semantic 3D data. SUNCG \parencite{Song_Yu_Zeng_Chang_Savva_Funkhouser_2017} offers a variety of indoor scenes with CAD-quality geometry and annotations. However, the level of detail and complexity of the different building elements in such crowd-sourced synthetic approaches is extremely limited when compared to 3D scanned alternatives. Synthetic datasets lack natural transformation and topological properties of objects in real-world settings. 

Furthermore, there is a broad body of literature focused on synthesizing indoor scenes by learning from prior data \parencite{zhang2019survey}. While such approaches are mainly focused on predicting furniture placements and arrangement in an empty \parencite{li2019grains,Fisher2012} or partially populated scene \parencite{kermani2016learning,keshavarzi2020scenegen}, they are also dependent on the quality and diversity of the input data in their training stage. Procedural models have also been widely used in generating full buildings \parencite{muller2006procedural,saldana2013procedural}, furniture layout \parencite{merrell2011interactive,germer2009procedural} and manipulating indoor scenes \parencite{Yu2011}. Yet again, the outputs of such methods lack the complexity of real-world captured data, falling short of being effectively utilized in common computer vision tasks.

Therefore, augmenting large scale datasets of 3D geometry which correspond to the complexity of the built environments is still an open challenge. Motivated by this challenge, we introduce GenScan, a generative system that populates synthetic 3D scan datasets. GenScan generates new semantic scanning datasets by transforming and re-texturing the existing 3D scanning data in a parametric fashion. The system takes an existing captured 3D scan as an input and outputs alternative variations of the building and furniture layout with manipulated texture maps. The process is fully automated and can also be manually controlled with a user in the loop. Such an approach results in the production of multiple data points from a single scan for 3D deep learning applications. 

\begin{figure}
  \centering
  \includegraphics[width=\columnwidth]{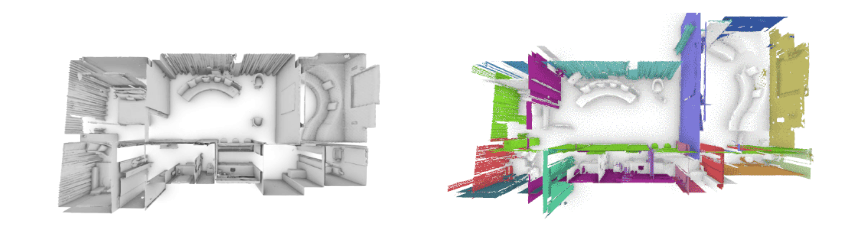}
  \caption{Results of the parametric modification (right) of an input scan (left)\label{fig:geomeodification}}
\end{figure}


\begin{figure*}
  \centering
  \includegraphics[width=\columnwidth]{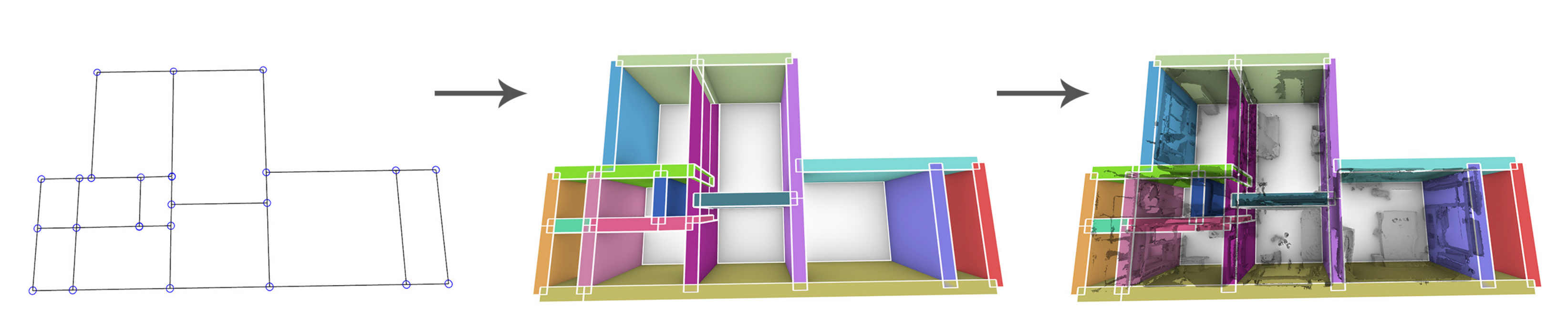}
  \caption{ Wall extraction module. We use the estimated floorplan layout and door sizes to construct threshold bounding boxes centered on each parametric line. With this method we classify wall elements (colored) and non-wall elements (white) in the scene.\label{fig:wallextraction}}
\end{figure*}

\section{Methodology}
The general workflow of the system consists of four main components. First, we predict the floorplan of the input 3D scan using a hybrid deep neural network (DNN). We classify what type of building the input model is and estimate what common finishing wall to wall distance the input model holds. Second, to avoid inconsistencies in the manipulated walls, we parameterize all generated vectors to prepare for element transformation. Third, we classify wall elements of the 3D scan using the predicted floorplan and automated thresholds, applying parametric transformations to all wall and non-wall elements separately. Finally, we apply a style transfer algorithm using a combination of a pre-trained VGG network and gradient descent module to current texture maps to generate new textures for the generate scenes. 

\begin{figure*}
  \centering
  \includegraphics[width=\columnwidth]{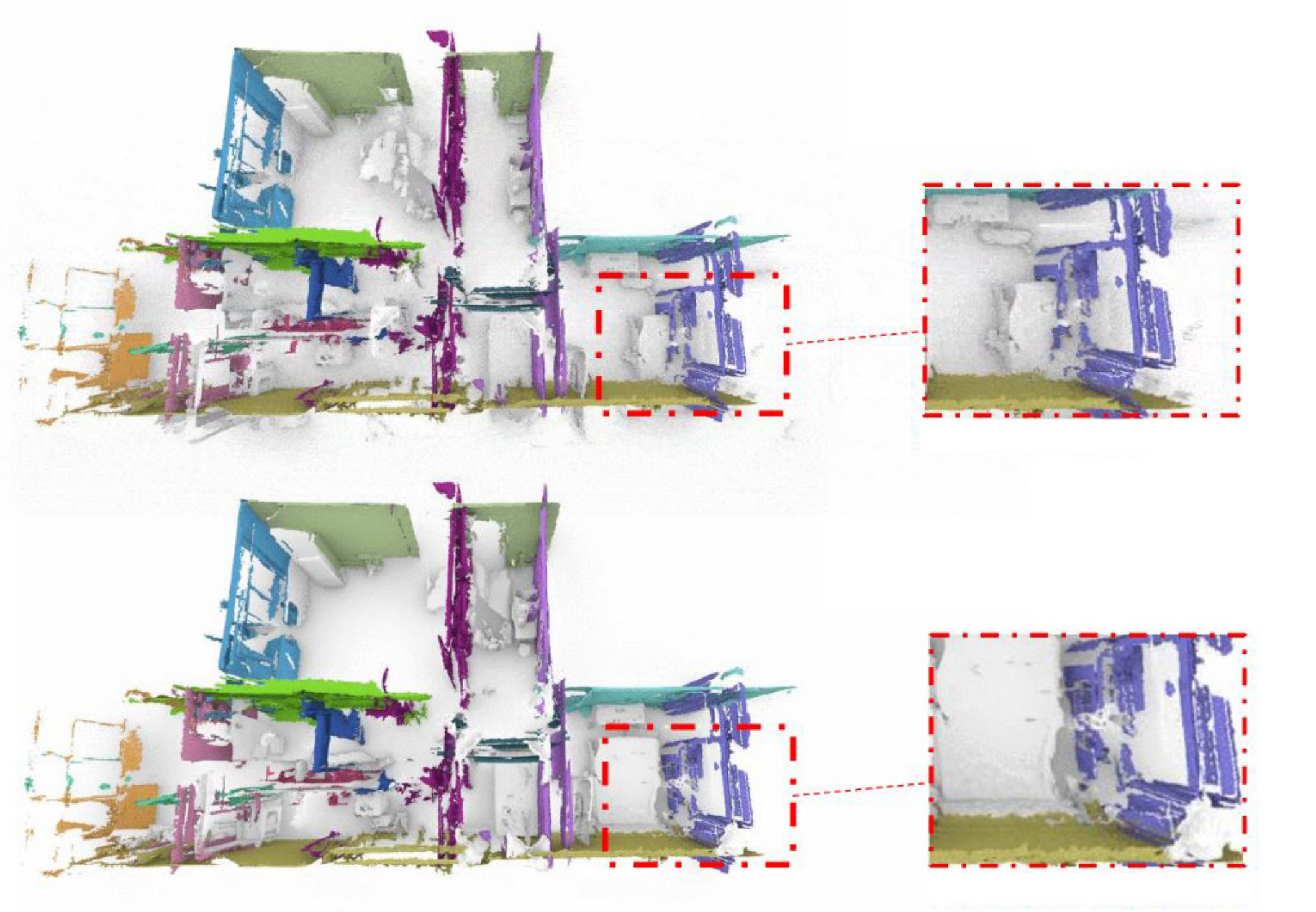}
  \caption{Transformation on wall elements only (top). Transformations on wall elements and closest furniture correspondingly\label{fig:conflictsolved}}
\end{figure*}

\subsection{Parameterization}
As shown in Figure ~\ref{fig:parameterizer}, moving an individual wall or a group wall with a certain transformation matrix produces inconsistency in the generated output layout, with unwanted gaps and voids emerging between corner points of the floorplan. We instead assign transformations to the corresponding nodes of the corner coordinates of the target wall elements. We utilize a modified implementation of \parencite{keshavarzi2020sketchopt, bergin2020automated} to parametrize the extracted floorplan. This would manipulate all lines connected to the transformed node. However, to avoid distortion of the orthogonal nature of the building floorplans, we merge co-linear paths that connect to each other with a mutual node and share the same direction vector. Next, we identify the array of nodes that are located on the co-linear lines. After applying transformations to the connected line node array, we construct new polylines from each node array. This would result in a fully automated parametric model that takes transformation vectors and connected line indices as an input and outputs a new floorplan layout without producing undesired gaps and floorplan voids

\subsection{Wall Extraction}
To classify the walls and movable edges of the input 3D scan, we use the original parametric model to extrude threshold bounding boxes centered on each of co-linear parametric lines generated in the previous step. We then construct a bounding box for each available mesh in the 3D scan input, and test if inscribes within any of the connected line bounding boxes. With this method, we estimate whether a mesh is part of the building wall system or not, and if so, we can find out which connected wall is it subscribed to. To define the width and threshold of the connected line bounding box, we take advantage of the extracted door sizes provided by the hybrid DNN module introduced in Liu et al's method \parencite{Liu_Wu_Furukawa_2018}. Based on the door sizes, we can classify what type of building the input model is and estimate what common finishing wall to wall distance the input model holds. This distance can later be verified by measuring the peak range in a vertical section histogram. However, the later verification is not always precise, as elements such as tall bookshelves and cabinets may interfere with thickness estimation of the walls.

\begin{figure*}
  \centering
  \includegraphics[width=\columnwidth]{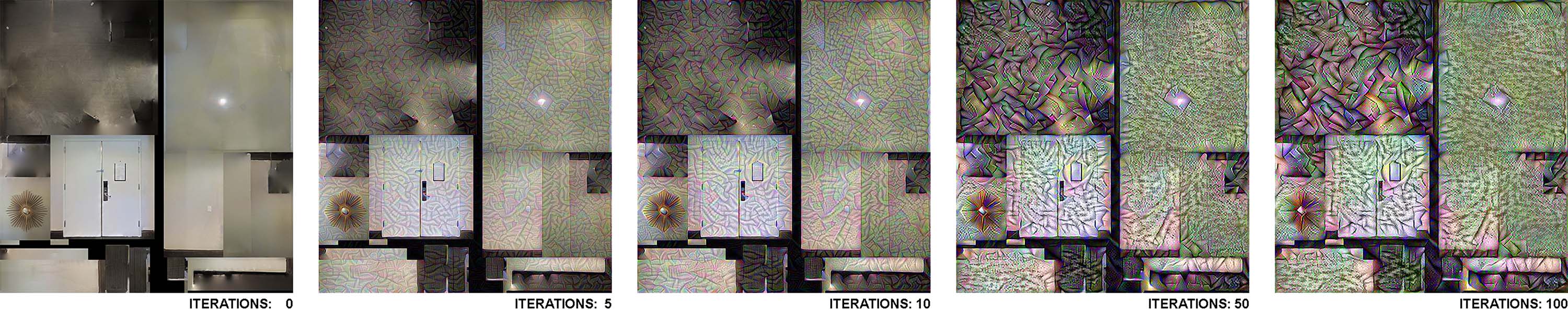}
  \caption{Iterations of the style transfer gradient descent algorithm.
  \label{fig:iterations}}
  
\end{figure*}

\subsection{Model Transformation}
Given a connected line index and an offset number, all nodes corresponding to the target line would be transformed in the direction perpendicular to the connected line. In many cases, this would not only affect the target line itself, but also change the size of neighboring connected lines. After all transformations are applied to the nodes of the graph, we calculate the difference between the transformation matrix of the initial geometry and the final geometry. This includes a two-dimensional translation vector defining the variance in the position and also a scale factor computed from the center of each line. Next, we apply the transformation vector of each connected line to all input meshes included in the corresponding bounding box. This would result in the parametric movement of the estimated walls, while maintaining the overall node graph constructed between all wall elements. By applying the scale transformation specifically to the x and y directions, we stretch and shrink the walls to avoid unwanted architectural inconsistencies and prevent the transformed output from containing irrelevant void and structural gaps.

However, as shown in Figure ~\ref{fig:wallextraction}, in many cases the modification made to the walls would overlap with non-wall elements or the building furniture. This would result in conflicting mesh artifacts in certain clusters. To address this problem, we calculate the center coordinates of each bounding box assigned to non-wall meshes and perform a closest point search with the parametric line system to find the closest wall. We then transform each mesh with the two-dimensional position translation vector of the corresponding closest wall, with a non-liner factor of its distance to the wall. Therefore, a non-wall mesh element closer to the wall would have a much similar transformation function to the wall itself, than a non-wall mesh element located in the middle of the room. This would allow furniture to move close and far in relation to each other, instead of moving in a similar direction altogether.

\begin{figure}
  \centering
  \includegraphics[width=0.6\columnwidth]{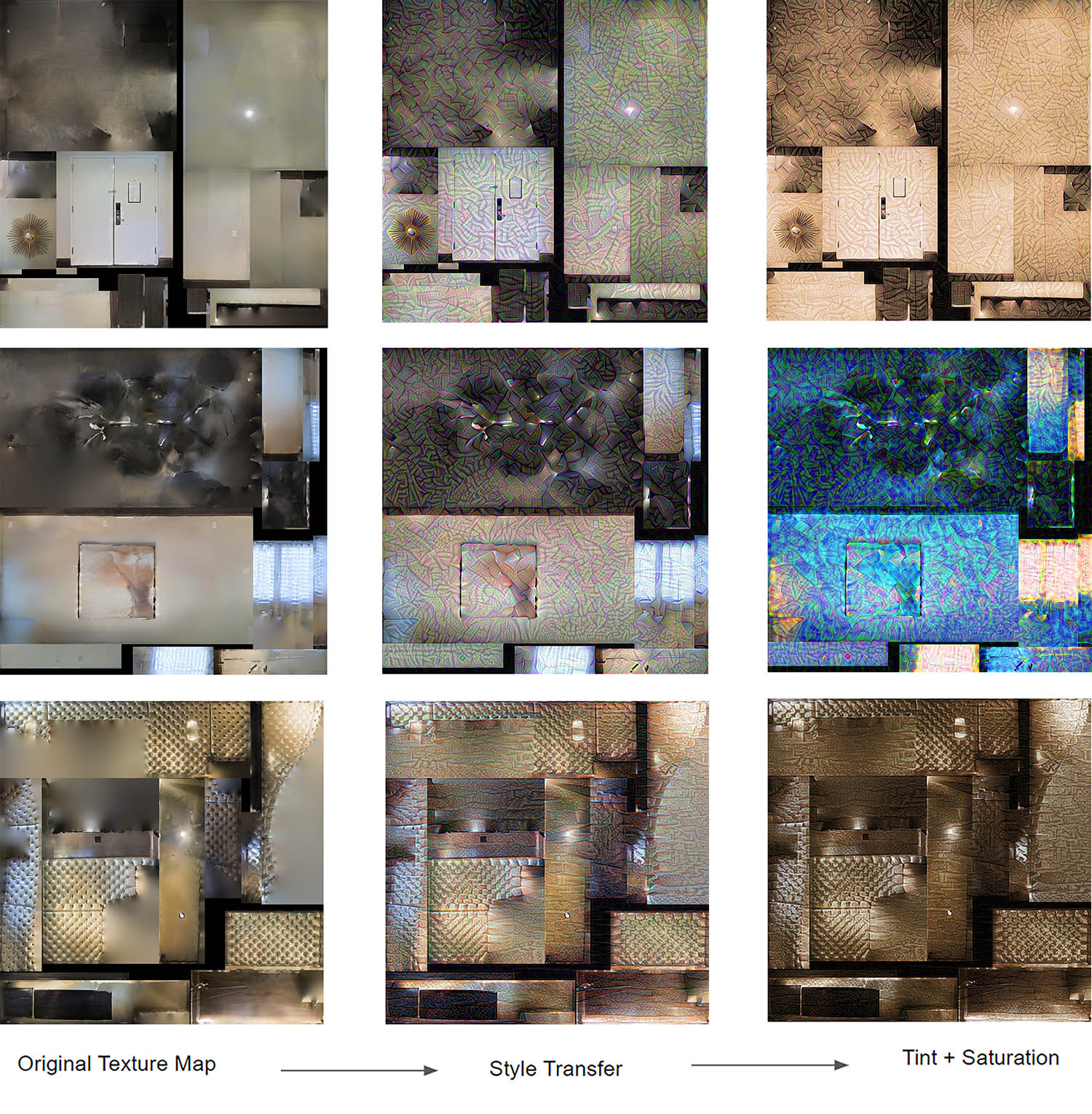}
  \caption{Different texture maps modified through style transfer and color modification. Permutations of matching style transfer with modified tints, hues, and saturation can be applied to generate diverse texture maps.
  \label{fig:hue}}
\end{figure}

\subsection{Model generation}

The parametric model can be modified to alternate layouts using two main approaches. First, by manually inputting the system a list of parametric line indices and a corresponding offset value, which requires a user in the loop. The second approach is by providing a random range of offsets values to be assigned to random parametric lines of the model. Such method, allows mass generations  of synthetic 3D scans which can be later filtered and sorted by implementing evaluation functions. Figure ~\ref{fig:conflictsolved} illustrates a random floorplan generation of 3D scan using this method.

\begin{figure*}
  \centering
  \includegraphics[width=\columnwidth]{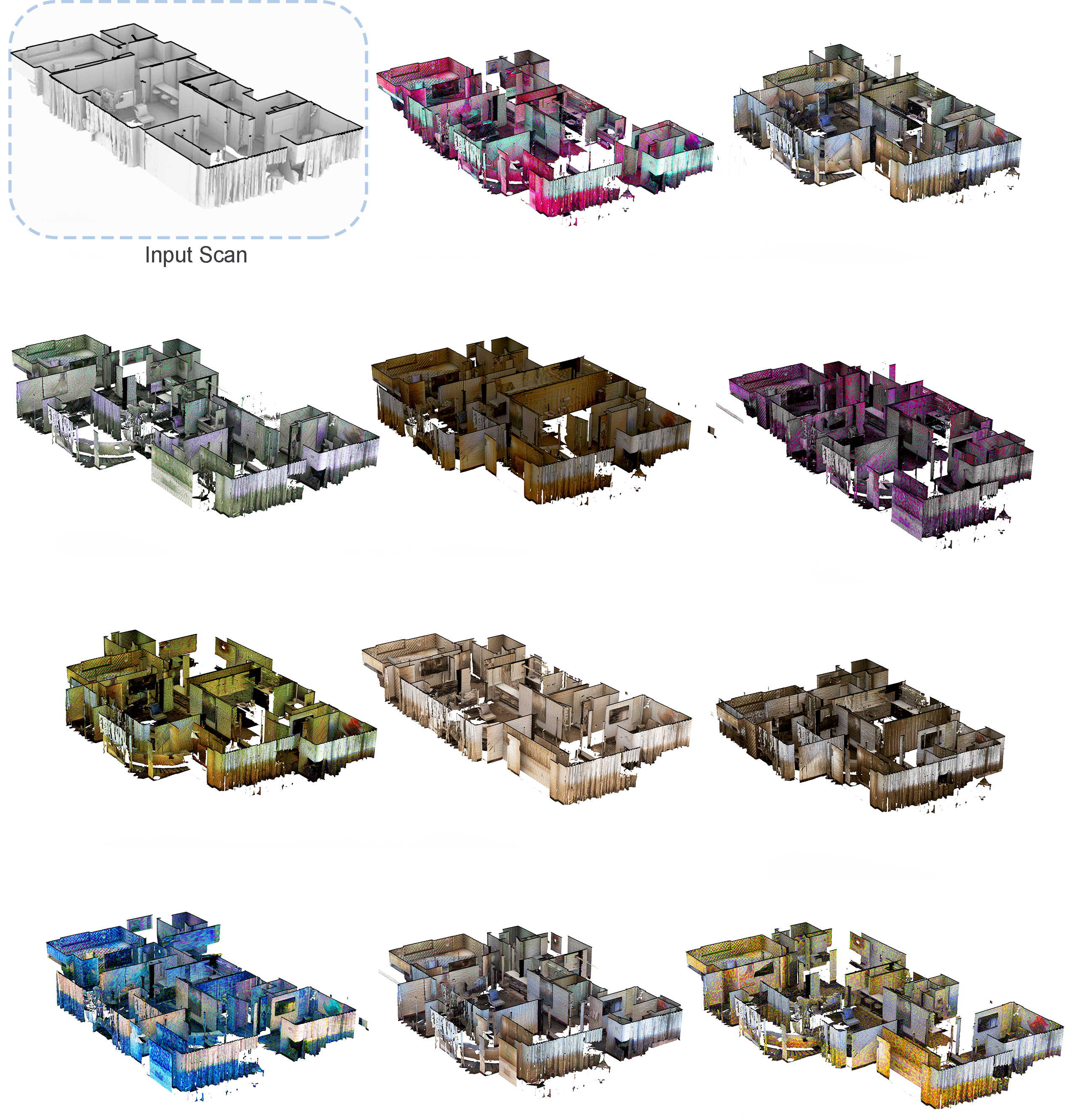}
  \caption{Examples of 3D mesh population from an input scan (top left) with modified floor geometries, texture elements, and colors..\label{fig:final}}
\end{figure*}

\section{Texture Generation}
After applying parameterized geometry transformations to the scanned data, we aim to change the overall visual appearance of the newly generated mesh by editing the associated texture maps. Within our texture modification pipeline, we follow two steps to modify the texture maps of the original mesh provided by the the input scan data. First, we take all the texture maps associated within one scan and apply a simple style transfer to each of the textures. Next, we take the generated texture map and apply corrections to its image characteristics such as hue, saturation, and tint, etc. Finally by updating the texture coordinates of the vertices in the newly generated geometry, we are able to match the style transferred texture maps accordingly.   

We implement the style transfer method introduced by Gatys et al. \parencite{Gatys2015}. We incorporate a pre-trained VGG network to output a style transferred texture map. We calculate the content loss and style loss of our generated image at each iteration of the algorithm and run a gradient descent module until we reach an iteration that looks visually convincing. In Figure ~\ref{fig:iterations}, we illustrate how the output image converges to the style image while the content loss and image loss are being minimized. The higher the number of iterations the more distinct the style is on the texture map, therefore, for a more subtle effect we choose a lower number of iterations for its realism. We apply the transfer technique to modify the texture maps included with the Matterport scans. Style transfer would allow diverse modifications of the input textures, an easy and efficient way to blend a generate variations within a single content texture.The style transfer implemented can be the same for each texture or unique. For example, each room or part of the mesh can have its own different texture modification. Our application of style transfer is to change our existing texture maps to look like new textures using this established technique. By using different style images we create rooms that look like they are made from brick, wood, or even a wallpaper laid on them. This versatility of style transfer allows the subset of data regeneration limitless and provides a unique enough new mesh that can be used for our original motivation.

Finally to allow for more texture variation and realism, we apply a post-processing module of hue, saturation, and tint adjustment to the texture maps. In Figure ~\ref{fig:hue}, we illustrate a variety of textures we can generate with control over these parameters. At the end of our pipeline, we use the original texture to adjust these parameters of the texture map image. We achieve this by converting the image into an RGBA array that we can shift and scale dictated by the desired effect. Overall, through just the texture modification process, we have control and access to infinite choices in style image and parameterization of image characteristics mentioned above. Figure ~\ref{fig:final} displays just a few of the possible final floor layouts created with GenScan.



\section{Discussions and Conclusion}
GenScan applies automated parameterization and texture modification of 3D scanned geometrical data to produce bootstrapped samples of 3D scanned data. Given data for just a single scan, GenScan actively produces valid synthetic geometric and textured data of multiple potential layouts resulting in floor plans with modified floor geometries, texture elements, and colors. We believe our system would allow for mass parametric augmentation to expand the currently limited 3D geometry datasets commonly used in 3D computer vision and deep learning tasks. Such an approach results in the production of multiple data points from a single scan for 3D deep learning methodologies. This methodology can have various impacts and applications across multiple industries including design optimization, computer vision, virtual and augmented reality, and construction applications. 

While the current GenScan system has the ability to parameterize walls and major building elements extracted from the floorplan layout, it does not cover parameterizing smaller room elements such as chairs, beds, tables, desk, etc. Such objects not only need to be identified using semantic segmentation methods, a parametric relationship would also need to be established to allow relevant layout modifications. Furthermore, generating non-orthogonal layouts and extend parameterization to distorted and curved layouts can be also considered as next steps to this study. Another limitation of our system lies in the inability to modify the textures of specific walls and non-wall objects of our choosing. Identifying specific areas of the texture maps to regenerate and filling in gaps produced by expanding layout would result in a cleaner 3D model. Moreover, applying unique changes to specific parts of the texture maps instead of the whole map would allow for even greater customization, variability, and realism of the data. Finally, streamlining our implementation of the texture modification process in our pipeline will achieve higher texture resolution quality in an efficient time period. 

\section{Acknowledgments}
We acknowledge the generous support from the following research grants: a FHL Vive Center for Enhanced Reality Seed Grant, and the Siemens Berkeley Industrial Partnership Grant, ONR N00014-19-1-2066. Furthermore, we would like to thank Woojin Ko and Avinash Nandakumar for their assistance in generating synthetic texture maps, and Chen Liu and Tommy Wei for their help in integrating and troubleshooting the FloorNet module within our system.

%
%
%
%
\printbibliography




\end{document}